\documentclass[pmlr,twocolumn,10pt]{jmlr} % W&CP article

\usepackage{booktabs}
\usepackage{siunitx}

\usepackage[switch]{lineno}

\theorembodyfont{\upshape}
\theoremheaderfont{\scshape}
\theorempostheader{:}
\theoremsep{\newline}

\jmlrvolume{Page}
\jmlryear{2024}
\jmlrsubmitted{}
\jmlrpublished{}
\jmlrworkshop{ArXiv Version: 2024, April 15th @ ArXiv} % W&CP title

\title[Self-Supervised Learning Featuring Small-Scale Image Dataset for Treatable Retinal Diseases Classification]{Self-Supervised Learning Featuring Small-Scale Image Dataset for Treatable Retinal Diseases Classification}

\author{%
\Name{Luffina C. Huang} \Email{luffina.c.huang@rice.edu}\\
\addr Rice University
\AND
\Name{Darren J. Chiu} \Email{djchiu@bu.edu}\\
\addr Boston University
\AND
\Name{Manish Mehta} \Email{manishmehta.work22@gmail.com}\\
\addr Microsoft
}

\begin{document}

\maketitle

\begin{abstract}

Automated medical diagnosis through image-based neural networks has increased in popularity and matured over years. Nevertheless, it is confined by the scarcity of medical images and the expensive labor annotation costs. Self-Supervised Learning (SSL) is an good alternative to Transfer Learning (TL) and is suitable for imbalanced image datasets. In this study, we assess four pretrained SSL models and two TL models in treatable retinal diseases classification using small-scale Optical Coherence Tomography (OCT) images ranging from 125 to 4000 with balanced or imbalanced distribution for training. The proposed SSL model achieves the state-of-art accuracy of 98.84\% using only 4,000 training images. Our results suggest the SSL models provide superior performance under both the balanced and imbalanced training scenarios. The SSL model with MoCo-v2 scheme has consistent good performance under the imbalanced scenario and, especially, surpasses the other models when the training set is less than 500 images. 

\end{abstract}


\paragraph*{Data Availability}

This paper uses the UCSD Dataset \citep{kermany2018identifying}, which is publicly available on the Mendeley Data repository \citep{kermany2018labeled}.

\section{Introduction}
\label{sec:intro}
Automated medical diagnosis through emerging digital technologies has been a popular field of study in recent years, enhancing precision diagnosis and the quality of medicine. Image-based machine learning and deep learning models are prevailing in this direction and retinal diseases classification is one of vital cases. 

Age-related macular degeneration (AMD) is a leading cause of irreversible vision loss and blindness in developed countries. In the U.S.A., it was estimated that 18.34 million (11.64\%) of Americans aged 40 and above had early-stage AMD, and 1.49 million (0.94\%) of the same age group had late-stage AMD which impaired their vision in 2019 \citep{rein2022prevalence}. Although drusen, the yellow/white deposits that accumulate under the retinal pigmented epithelium, are the early clinical sign of AMD, they often precede and increase the chance of choroidal neovascularization (CNV), the later stage of AMD that jeopardizes vision. Diabetic eye disease is a major global cause of visual impairment. Among the estimated 285 million people with diabetes mellitus globally, about 11\% have vision-threatening conditions, such as diabetic macular edema (DME) , which involve accumulated fluid in the central retina and increases the risk of visual loss if left untreated \citep{lee2015epidemiology}. 

\begin{figure}[t]
\vskip 0.0in
\begin{center}
\centerline{\includegraphics[width=1 \columnwidth]{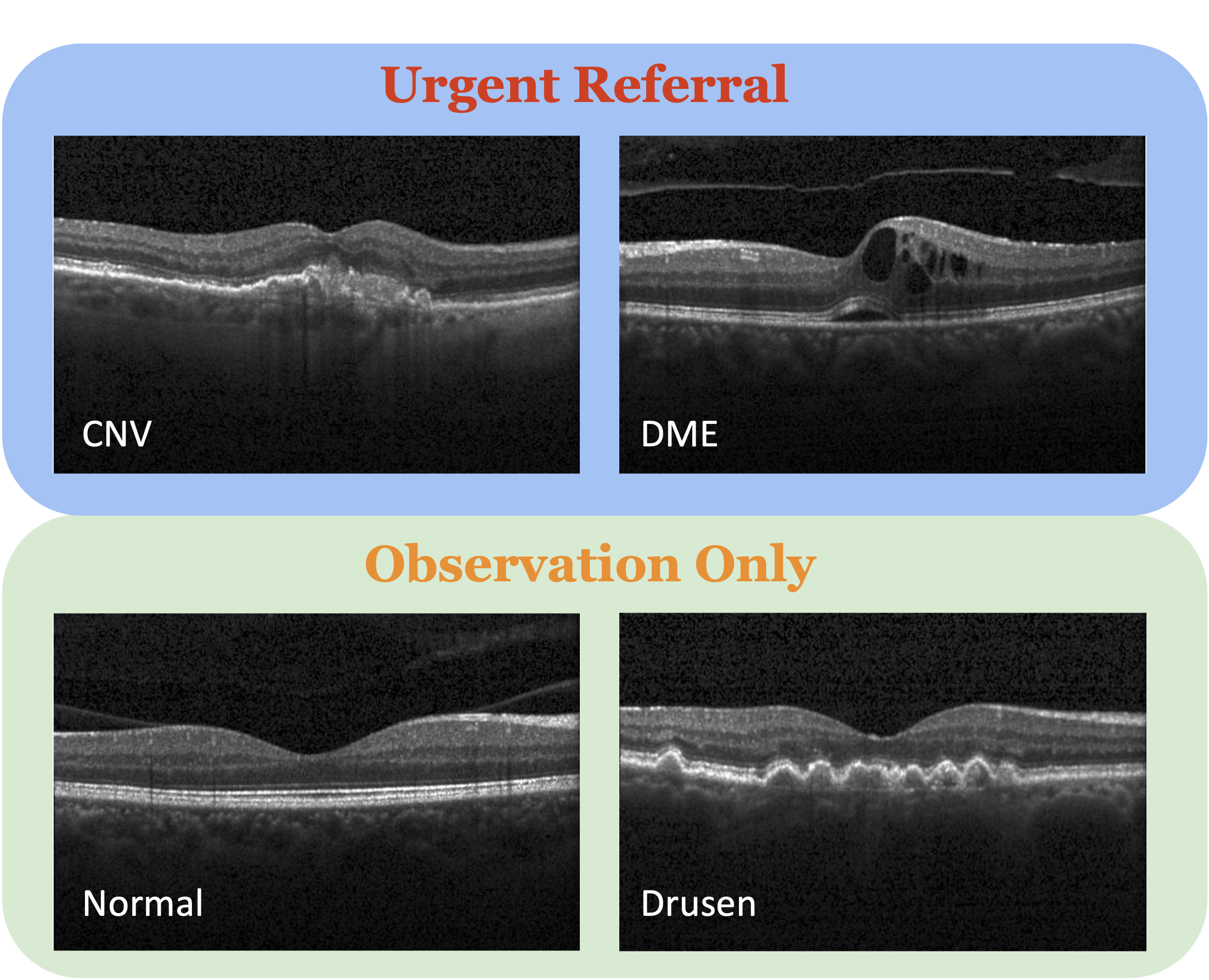}}
\end{center}
\vskip -0.3in
\caption{Optical coherence tomography raw images from UCSD Dataset. Choroidal neovascularization (CNV) and Diabetic macular edema (DME) are categoried into urgent referral, while Drusen and Normal are categoried into observation only. }
\label{oct_raw_imag}
\end{figure}

Optical coherence tomography (OCT) imaging provides a clear cross-sectional visualization of individual layers of retina. OCT is critical to guide the diagnosis and treatment of AMD and DME \citep{boned2021optical}, as shown in Figure \ref{oct_raw_imag}. Classification of OCT images could facilitate early detection, disease progress monitoring, and improve the prognosis prediction. Early intervention could avoid the severe visual acuity losses. Given the high prevalence of AMD and DME, a reliable automatic classification of OCT images could help alleviate the considerable burden of screening and monitoring on the health providers.

Some studies have explored machine learning and statistics methods on optical coherence tomography dataset for various diagnosis purposes.  \citet{farsiu2014quantitative} trains a generalized linear regression model using manually engineered features of the retinal pigment epithelium layer to classify the AMD and normal OCT images. \cite{srinivasan2014fully} uses histogram of oriented gradients (HOG) descriptors and SVMs to classify the spectral domain-OCT volume as normal, AMD, or DME. However, the performance of these methods rely on feature extraction processes, which require domain expertise and are labor intense.

The integration of deep learning into medical images has the potential to revolutionize disease diagnosis, risk stratification and treatment planning in various medical domains, such as retinal disease, and respiratory disease \citep{aggarwal2021diagnostic, degrave2021ai}. Quick and reliable identification of suspicious lesions can reduce the workload of healthcare professionals and help improve the quality of healthcare. The successful adoption of Deep Learning technology into clinical workflows is contingent upon achieving diagnostic accuracy that is at least on par with that of healthcare professionals. However, deep learning requires a large-scale labeled training dataset, which limits the application of deep learning to medical domains due to the scarcity of medical images and the expensive annotation costs. \cite{8120662, saleh2021transfer, arefin2021non, andreeva2021dr}. 

Transfer learning is a method which pre-trains a model with classical supervised learning on a large, unrelated dataset. This pre-trained model then undergoes so-called “fine-tuning” based on the data in the training dataset. However, large datasets of manually labeled example images are expensive to compile, considerably limiting the utility of this method in many cases. Transfer learning aims to address the challenge of limited data in a given domain by leveraging prior knowledge of pre-learning feature maps \citep{kermany2018identifying, chan2017earlier}. However, there is a potential risk that the representations learned during pre-training will not fully adapt to the downstream task \citep{saleh2021transfer}. Furthermore, the medical images dataset are usually imbalanced with the minority class being of main interest. This will lead to the transfer learning model focusing on learning the majority classes and failing to generalize to predict the minority class of interest. 

Self-supervised learning (SSL) bridges the gap between supervised learning and unsupervised learning. Self-supervised learning is able to pre-train models on unlabelled data using “pretext” tasks (data augmentation) as an alternative way to extract useful representations before tuning them for the downstream target task. Through solving the pretext tasks, pseudo labels are generated automatically based on the dataset’s properties \citep{wang2023review}. Self-supervised learning has demonstrated similar potential to transfer learning in many tasks such as object localization, speech representation and image classification. Furthermore, self-supervised learning models have been reported to handle the imbalanced dataset better than the transfer learning models \citep{yang2020transfer, yang2020rethinking}. Some studies have employed transfer learning for OCT images classification \citep{kermany2018identifying}. However, self-supervised learning has not been explored on OCT imaging yet, especially for small-scale size datasets. Hence, there is an unmet need to solve issues of particularly few manually-labeled examples available in the regime of medical image classification \citep{wang2023review}. 

In this study, we leverage four self-supervised learning pre-training models, Swapping Assignments between Multiple Views (SwAV) and Momentum Contrast (MoCo v2), Simple Framework for Contrastive Learning of Visual Representations (SimCLR), and Bootstrap Your Own Latent (BYOL) for the automated diagnosis of retinal disease in the UCSD Dataset, compared to two transfer learning models, Inception-v3 and ResNet50. By applying those models into small subsets of the UCSD dataset, we aim to determine which models are favorable for regimes with low data availability and data imbalance in OCT images. This framework could be further generalized and hopefully increase availability of automated medical diagnosis models for other treatable diseases which have low training medical images. The contributions of this article are as the follow:
\begin{enumerate}
\item We explore a self-supervised learning framework on the UCSD Dataset for retinal diseases classification achieving state-of-art performance featuring small-scale OCT images, encouraging the usage of self-supervised learning models for automated medical diagnosis. 
\item We assess the performance of self-supervised learning models on imbalanced and very small OCT dataset (n$<=$500). Results show that MoCo-v2 scheme has promising outcomes.
\item The proposed framework takes very short time to complete one training task (only 1 hour for training 4,000 OCT images), indicating computational efficiency for pre-trained self-supervised learning models. 
\end{enumerate}

\section{Related Works}

\subsection{Automated medical diagnosis on OCT}

Automated diagnosis from OCT images can be achieved by two main approaches: feature engineering and deep learning. 
Feature engineering methods extract the pathological biomarkers from the OCT images and use machine learning classifiers to make the diagnosis decision. Previous studies have employed morphological features, such as the segmentation, volumes, area and thickness of different retinal layers, and computer vision algorithm-extracted features, such as histogram of oriented gradients (HOG), patch detection and bag of words (BoW), to encode disease-specific information from the OCT images. Various machine learning classifiers, such as the support vector machine, Bayesian classifier and random forest classifier, have been used to obtain the medical diagnosis from the extracted features \citep{farsiu2014quantitative, srinivasan2014fully, venhuizen2017automated}. However, feature engineering usually requires manual annotation, domain expertise and strong image-preprocessing techniques to achieve high performance. 

Deep learning with convolutional neural networks (CNN) has attracted more attention in the field of automated medical images diagnosis, including OCT images. Previous studies have applied transfer learning with CNN models pre-trained on a large, unrelated dataset in a supervised learning manner for OCT image classification. \citet{kermany2018identifying} used Inception-v3 pretrained on the ImageNet dataset \citep{szegedy2016rethinking} for the classification of OCT images into four classes: CNV, DME, drusen, and normal, achieving a classification accuracy of 96.6\%. They also demonstrated the generalizability of their approach for pneumonia diagnosis using chest X-ray images. \citet{serener2019dry} fine-tuned the Residual Net (ResNet) pretrained on ImageNet to classify the early and late stage AMD, and showed that it outperformed AlexNet. Furthermore, \citep{sotoudeh2022multi} proposed multi-scale frameworks based on the feature pyramid network (FPN) structure using pre-trained ImageNet weights are proposed for the classification of the OCT images. Some studies have trained the CNN models from scratch using OCT images without transferring knowledge from pretrained models. For instance, \citet{rastogi2019detection} trained a Densely Connected Convolutional Neural Network (DenseNet) on OCT images in the UCSD Dataset and achieved an accuracy of 97.65\%. \citet{paul2020octx} proposed a Deep Ensemble Network that fused the output from three architectures (DenseNet, VGG16, Inception-v3) and a fourth custom CNN trained on OCT images to further improve the performance.

\subsection{Self-supervised learning for visual tasks and medical images}

Self-supervised learning uses data augmentation techniques to create the views of the samples to distinguish between positive and negative samples using contrastive loss, where positive samples are the augmented views of a target sample, while negative samples are from other non-target samples within the training set \citep{MIR-2022-08-271}. This allows pre-learning representations on unlabeled data before fine-tuning them for downstream target tasks with labeled data and has shown perform on par with their supervised counterparts \citep{yang2020transfer}. 

SimCLR \citep{chen2020simple} uses an end-to-end learning architecture and learns representations by maximizing the agreement between different augmented views of the same sample via a contrastive loss calculation for each batch. Given the numbers of negative samples in a training batch have a significant impact on the accuracy, this approach requires the use of very large batch sizes, strong data augmentation techniques, and extensive training epochs to achieve high performance. 

To decrease the dependence on the batch size, which is limited by GPU memory, MoCo \citep{he2020momentum} proposes a dictionary-like first-in-first-out queue to progressively replace the negative samples in this queue batch by batch. MoCo utilizes a momentum encoder to update the parameters for each label in the queue and keeps a large and consistent dictionary that enhances the performance of the contrastive learning . MoCo-v2 \citep{chen2020improved} further enhance the performance by adding a multiple layer projection head, strong data-augmented view, and a cosine learning rate schedule. 

BYOL \citep{grill2020bootstrap} utilizes two networks: an online network and a target network. The online network receives one view of the data, while the target network receives another view of the same data. The loss function is the mean square error between the two output projections. In contrast to SimCLR and MoCo, which depend on many negative pairs for learning the distinctive features, BYOL aligns data-augmented views within positive pairs, but without using negative pairs.	
					
Instead of learning the instance-based discrimination mentioned above, SwAV \citep{NEURIPS2020_70feb62b} utilizes the cluster assignments to learn the representations exploiting cluster-based discrimination batch by batch. The key idea of SwAV is to assign data-augmented views of the same data sample to the same cluster using a “swapped” prediction technique.  This approach can work with large and small batch sizes without the need of a momentum encoder or a large memory bank.

Self-supervised learning has gained more attention in medical applications. Previous studies have employed self-supervised pretext tasks to perform image reconstruction tasks on corrupted or incomplete inputs, such as retinal images \citep{hervella2018retinal}.  More recently, MoCo pretraining and training has been applied to chest-x-ray images. For instance, \citet{sriram2021covid} applied MoCo to predict adverse events and oxygen requirements for COVID-19 patients from single or multiple chest X-ray images. \citet{sowrirajan2021moco} proposed MoCo-CXR, a variant of MoCo that adapts the data augmentation strategy to chest X-ray images, and demonstrated its effectiveness for detecting various pathologies. They reported that linear models trained on the representations learned by MoCo-CXR outperformed those trained on random or ImageNet-pretrained representations for pleural effusion detection.

\section{Methods}

\subsection{Model Training Framework}

\begin{figure}[t]
\vskip 0.0in
\begin{center}
\centerline{\includegraphics[width=1 \columnwidth]{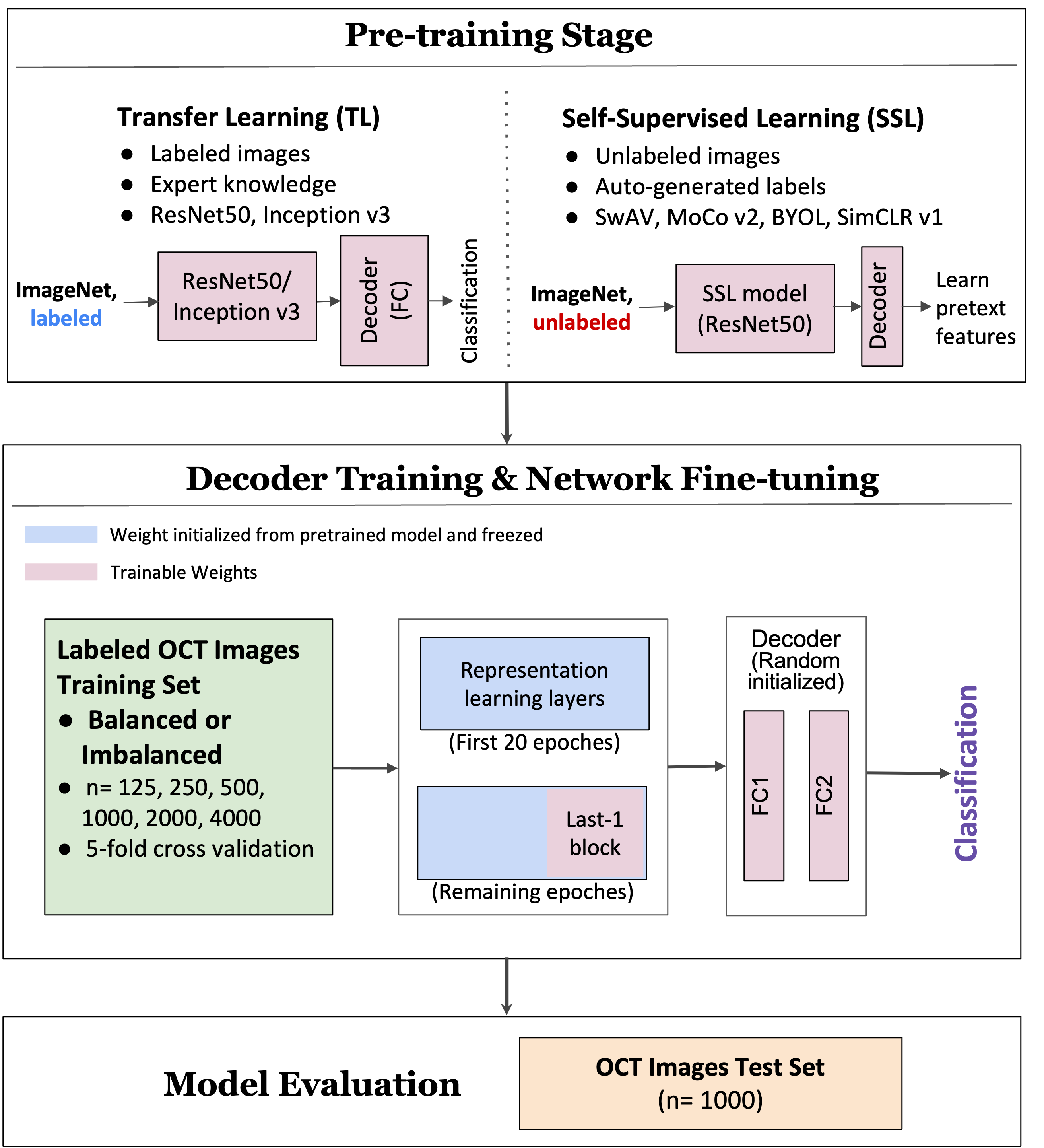}}
\end{center}
\vskip -0.3in
\caption{Overview of the pre-trained transfer learning and self-supervised learning framework, including pre-training stage, fine-tuning stage and final model evaluation.}
\label{Framework_Overview}
\end{figure}

We evaluate two transfer learning models and four self-supervised learning models, as shown in the Figure \ref{Framework_Overview}. For transfer learning, we used the ResNet50 and Inception-v3 supervised pre-trained on the labeled ImageNet data. The four self-supervised models are pre-trained on unlabelled ImageNet through SwAV, MoCo-v2, SimCLR-v1, and BYOL schemes, respectively. All of the four self-supervised pre-training use ResNet50 as backbone. As for the decoder training and network fine-tuning stage, the weights for representation layers are initialized as those pre-trained models. Two fully connected layers with 256 and 4 units are added as the decoder for all the six models. For the first 20 epochs, the weights of the whole representation layers are set as non-trainable, and only the decoder layers are trained. After the end of the first 20 epochs, for the remaining training epochs, we unfreeze the weights of the last block of representation layers to fine-tune the whole network. Note that, in order to support computational efficiency, the minimum and maximum training epochs are 50 and 100 with early stopping criteria of 10 epochs. 

\subsection{Implementation Details}

The training procedure involved processing batches of 128 images and updating the model weights using the Adam optimizer with a learning rate of 0.005. The other parameters of the optimizer, such as the weight decay, are set to the default values. The learning rate is warmed up and decayed by the annealed cosine learning rate schedule, which uses a higher learning rate to help the randomly-initialized fully-connected layers search the parameter space better in the first 20 epochs and at a much lower learning rate when the last one block of the representation layers are unfrozen.

The standard cross-entropy loss function is used as the objective function during training the balanced datasets. \citet{fernando2021dynamically} shows weighted cross entropy performs better than standard cross entropy and focal loss in various class imbalanced learning tasks using CNN models. Therefore, the weighted cross entropy loss is used when training the imbalanced datasets. The class weight is defined as ~(\ref{classweight}):

\begin{equation}
            \displaystyle
Classweight_{i} = \frac{\sum N_i}{N_i}
\label{classweight}
\end{equation}

where $N_i$ is the number of the samples for each class. For each model and dataset size condition, we apply 5-fold cross validation. During the training process, the validation losses and accuracies are monitored for each epoch to evaluate the model performance and generalization ability for the balanced training sets, whereas validation weighted F1 score and weighted losses are monitored for the imbalanced training sets \citep{bekkar2013evaluation}.

\subsection{Model Evaluation}
We used accuracy and weighted F1 calculated on the test set as the model evaluation metrics. The accuracy metric is defined as ~(\ref{accuracy}), the ratio of the number of correctly classified images to the total number of images in the test set: 
\begin{equation}
Accuracy = \frac{1}{N}\sum_{i=1}^n  1 (\hat{y_i} = {y_i})
\label{accuracy}
\end{equation}

where $\hat{y_i}$ is the predicted label of the $i$-th sample, ${y_i}$ is the corresponding true label and $N$ is the total numbers of images in dataset.\\
For each class, the F1 score is defined as ~(\ref{F1}) :
\begin{equation}
F1 score_{i} = \frac{2 * TP_i}{2*TP_i + FN_i + FP_i}
\label{F1}
\end{equation}

where $TP_i$ is the number of true positives, $FN_i$ is the number of false negatives, and $FP_i$ is the number of false positives for each class. \\
The weighted F1 is defined as ~(\ref{weighted_F1}):

\begin{equation}
            \displaystyle
Weighted F1 = \sum_{i=1}^4 \frac{N_i}{N} * F1score_{i}
\label{weighted_F1}
\end{equation}

where $N_i$ is the number of samples for each class.

\begin{figure*}[h]
\vskip 0.0in
\begin{center}
\centerline{\includegraphics[width=2 \columnwidth]{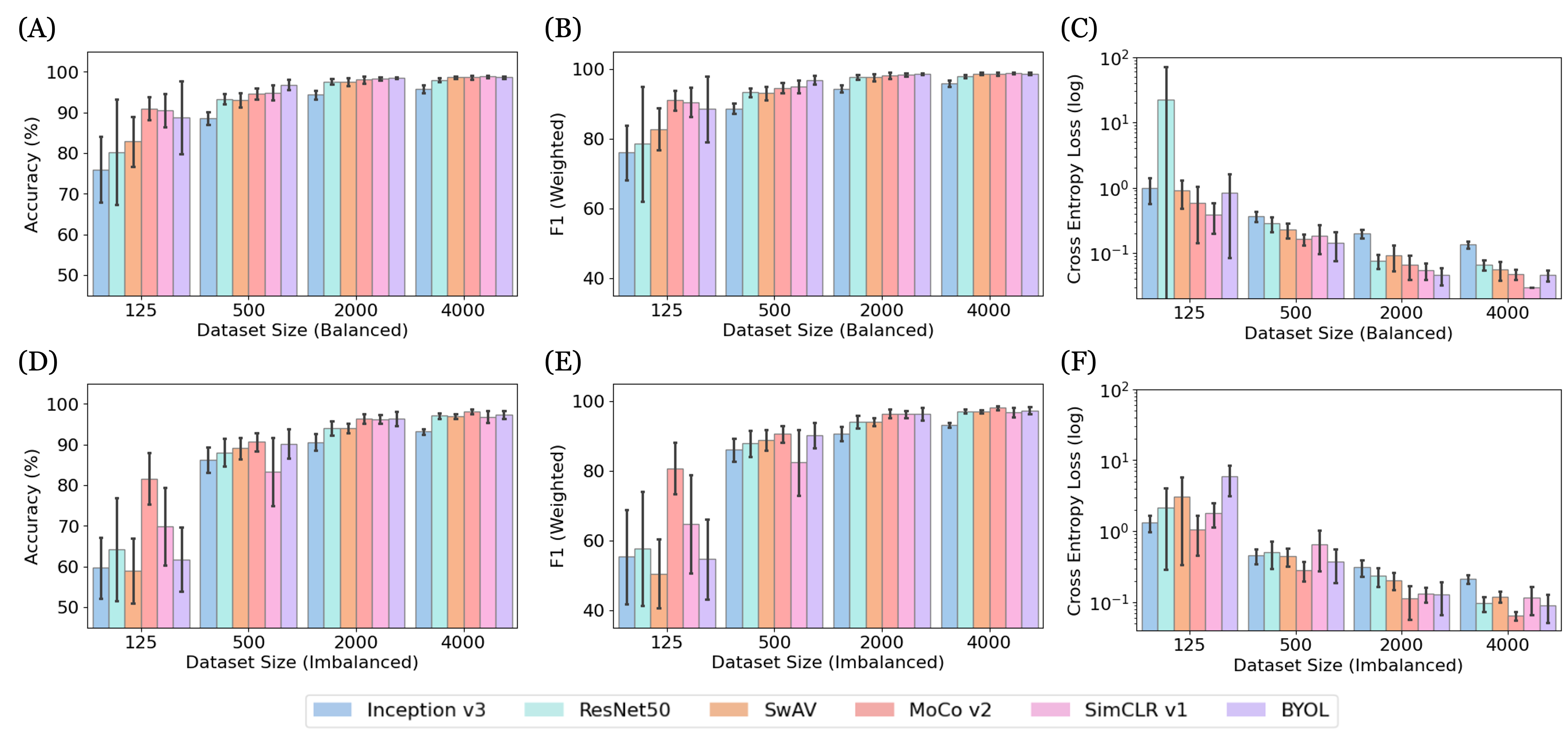}}
\end{center}
\vskip -0.3in
\caption{Bar plots presenting the mean and standard deviation (error bar) for accuracy, F1 score and cross entropy of six models trained on different sizes of dataset. Inception v3 and ResNet50 are transfer learning pre-trained models, while SwAV, Moco v2 and SimCLR v1 and BYOL are self-supervised learning pre-trained models. Plots in the upper panel are results from balanced dataset. Plots in the lower panel are results from imbalanced dataset.} 
\label{barplot}
\end{figure*}

\section{Dataset and Data Preprocessing}

\subsection{Dataset}

We use OCT images from the publicly available dataset, called UCSD Dataset \citep{kermany2018labeled}. The dataset consists of 108,312 OCT images in the training set and 1,000 OCT images in the test set. The training dataset has 37,206 CNV, 11,349 DME, 8,617 Drusen, and 51,140 Normal images. Images with CNV and DME are designated as 'Urgent Referral', which would demand urgent referral to an ophthalmologist for treatment. Images with drusen are designated as 'Routine Referrals'. Normal images were labeled for 'Observation'. 

\subsubsection{Training Dataset}

To examine the effectiveness of the various training methods under different scenarios of balanced or imbalanced training dataset, we generated two customized training sets of 4000 images from the training set of UCSD Dataset. First, the balanced dataset, where 1000 images are randomly selected from each class. Second, the imbalanced dataset, where the images were randomly selected according to the original data distribution. We also evaluate the performance based on the different sizes of the training set by selecting random subsets of the corresponding training set with sizes of 125, 250, 500, 1000, 2000 and 4000. To estimate the variability of each scenario, we applied 5-fold cross validation to calculate the standard deviation for each experiment condition.

\subsubsection{Testing Dataset}

The test dataset contained 250 images from each class and is independent of the patients in the training set. We used the entire test set to evaluate the performance of each fold of the trained model.

\begin{table*}[h]
\floatconts
  {tab:table1}%
  {\caption{Performance of six pre-trained models on the balanced training sub-datasets. The sub-dataset has the same numbers of OCT images among all categories. Comparison of results of transfer learning and self-supervised learning models. Data are presented as mean $\pm$ standard deviation.}}
{ \vskip -0.1in
\resizebox{0.9\linewidth}{!}{%
\begin{tabular}{{l}*{7}{c}}
\toprule
& & \multicolumn{2}{c}{Transfer Learning} & \multicolumn{4}{c}{Self-Supervised Learning} \\ 
\cmidrule(lr){3-4} \cmidrule(lr){5-8}
     Metric &  Dataset Size & Inception-v3 &    ResNet-50 &         SwAV &      MoCo-v2 &    SimCLR-v1 &         BYOL \\
\midrule
 & 125 &  75.96 $\pm$ 7.20 &  80.30 $\pm$ 11.60 & 82.88 $\pm$ 5.46 & \textbf{90.94} $\pm$ 2.51 & 90.54 $\pm$ 3.66 & 88.78 $\pm$ 7.98 \\
 & 250 & 84.44 $\pm$ 3.89 &  88.32 $\pm$ 3.40 & 91.86 $\pm$ 1.68 & \textbf{94.10} $\pm$ 2.13 & 92.08 $\pm$ 3.16 & 92.66 $\pm$ 2.55 \\
Accuracy & 500 & 88.56 $\pm$ 1.35 &  93.30 $\pm$ 1.11 & 93.06 $\pm$ 1.65 & 94.60 $\pm$ 1.26 & 94.88 $\pm$ 1.66 & \textbf{96.82} $\pm$ 1.15 \\
$\pm$SD (\%) & 1000 &  93.30 $\pm$ 0.19 &  96.30 $\pm$ 0.61 &  96.70 $\pm$ 0.73 &  96.14 $\pm$ 1.16 &  96.62 $\pm$ 0.37 &  \textbf{97.30} $\pm$ 0.47 \\
 & 2000 & 94.34 $\pm$ 0.99 & 97.64 $\pm$ 0.53 & 97.60 $\pm$ 0.87 & 98.08 $\pm$ 0.79 & 98.34 $\pm$ 0.32 & \textbf{98.52} $\pm$ 0.25 \\
 & 4000 &  95.80 $\pm$ 0.79 & 97.96 $\pm$ 0.41 & 98.62 $\pm$ 0.28 & 98.62 $\pm$ 0.39 & \textbf{98.84} $\pm$ 0.27 & 98.62 $\pm$ 0.30 \\
   \hline & 125 &  75.98 $\pm$ 7.06 &  78.48 $\pm$ 14.82 &  82.76 $\pm$ 5.42 &  \textbf{90.96} $\pm$ 2.51 &   90.49 $\pm$ 3.73 &  88.50 $\pm$ 8.39 \\
   & 250 &  84.18 $\pm$ 4.41 &  88.19 $\pm$ 3.71 &  91.85 $\pm$ 1.70 &  \textbf{94.12} $\pm$ 2.08 &  92.10 $\pm$ 3.13 &  92.63 $\pm$ 2.56 \\
  F1$\pm$SD (\%) & 500 &  88.60 $\pm$ 1.33 &  93.30  $\pm$ 1.11 &  93.03  $\pm$ 1.67 & 94.59 $\pm$ 1.27 &  94.89 $\pm$ 1.65 &  \textbf{96.82} $\pm$ 1.15 \\
   & 1000 &     93.31 $\pm$ 0.19 &  96.29 $\pm$ 0.61 &  96.70 $\pm$ 0.73 &  96.13 $\pm$ 1.16 &  96.62 $\pm$ 0.37 &     \textbf{97.30} $\pm$ 0.47 \\
   & 2000 &  94.34 $\pm$ 0.98 &  97.64 $\pm$ 0.53 &  97.60 $\pm$ 0.87 &  98.08 $\pm$ 0.79 &   98.34 $\pm$ 0.32 &     \textbf{98.52} $\pm$ 0.25 \\
   & 4000 &  95.80 $\pm$ 0.79 &  97.96 $\pm$ 0.41 & 98.62 $\pm$ 0.28 &  98.62 $\pm$ 0.39 &     \textbf{98.84} $\pm$ 0.26 &     98.62 $\pm$ 0.30 \\
 \hline & 125 &  0.99 $\pm$ 0.38 &  22.58 $\pm$ 43.20 &  0.90 $\pm$ 0.37 &  0.59 $\pm$ 0.40 &  \textbf{0.39} $\pm$ 0.17 &  0.85 $\pm$ 0.68 \\
 & 250 &  0.60 $\pm$ 0.25 &  0.79 $\pm$ 0.81 &  0.35 $\pm$ 0.12 &  \textbf{0.25} $\pm$ 0.20 &  0.32 $\pm$ 0.10 &  0.30 $\pm$ 0.12 \\
Loss$\pm$SD & 500 &  0.37 $\pm$ 0.06 &  0.28 $\pm$ 0.07 &  0.23 $\pm$ 0.06 &  0.16 $\pm$ 0.03 &  0.18 $\pm$ 0.07 &  \textbf{0.14} $\pm$ 0.06 \\
 & 1000 &  0.22 $\pm$ 0.01 &  0.12 $\pm$ 0.02 &  0.11 $\pm$ 0.03 &  0.11 $\pm$ 0.03 &  0.12 $\pm$ 0.03 &  \textbf{0.08} $\pm$ 0.02 \\
 & 2000 &  0.20 $\pm$ 0.03 &  0.07 $\pm$ 0.01 &  0.09 $\pm$ 0.03 &  0.07 $\pm$ 0.02 &  \textbf{0.05} $\pm$ 0.01 &  \textbf{0.05} $\pm$ 0.01 \\
 & 4000 &  0.14 $\pm$ 0.01 &  0.07 $\pm$ 0.01 &  0.06 $\pm$ 0.02 &  0.05 $\pm$ 0.01 &  \textbf{0.03} $\pm$ 0.00 &  0.05 $\pm$ 0.01 \\
\bottomrule
\end{tabular}%
} }
\end{table*}

\subsection{Dataset Augmentation and Preprocessing}

We further utilized several data augmentation techniques to increase the variation and number of training images, and prevent the overfitting problem. This encourages invariance of feature maps to nuisance variables such as rotation, zoom power, brightness, noise level, and aspect ratio.  Firstly, the training samples are transformed to have random rotations up to 30 degrees, random horizontal fip, and random crop, followed by resizing to the common 244x244 input size for models using ResNet50 as backbone or 299x299 input size for Inception-v3. 

These transformations generally behave as a regularization step, and prevent overfitting to the train set. The images in the test set are directly resized without any rotation, flipping, or cropping. Note that we only apply these transformations to 20\% of the training images. Secondly, the image tensors are normalized according to the true pixel brightness mean and standard deviation of the dataset set. Since the images are grayscale, the same normalization statistics are used for each RGB channel.

\begin{table*}[h]
\floatconts
  {tab:table2}%
  {\caption{Performance of six pre-trained models on the imbalanced training sub-datasets. The sub-dataset has the uneven distributions of OCT images among all categories. Comparison of results of transfer learning and self-supervised learning models. Data are presented as mean $\pm$ standard deviation.}}
{\vskip -0.1in
\resizebox{0.9\linewidth}{!}{
\begin{tabular}{{l}*{7}{c}}
\toprule
& & \multicolumn{2}{c}{Transfer Learning} & \multicolumn{4}{c}{Self-Supervised Learning} \\ 
\cmidrule(lr){3-4} \cmidrule(lr){5-8}
     Metric &  Dataset Size & Inception-v3 &    ResNet-50 &         SwAV &      MoCo-v2 &    SimCLR-v1 &         BYOL \\
\midrule
 & 125 &  59.68 $\pm$ 6.77 &  64.24 $\pm$ 11.26 & 58.96 $\pm$ 7.07 & \textbf{81.60} $\pm$ 5.68 & 69.86 $\pm$ 8.60 & 61.76 $\pm$ 7.03 \\
 & 250 & 79.06 $\pm$ 4.66 &  87.42 $\pm$ 2.53 & 78.36 $\pm$ 6.67 & \textbf{90.14} $\pm$ 2.37 & 82.52 $\pm$ 2.55 & 82.50 $\pm$ 6.43 \\
Accuracy & 500 & 86.22 $\pm$ 2.74 &  88.04 $\pm$ 3.05 & 89.10 $\pm$ 2.32 & \textbf{90.64} $\pm$ 2.02 & 83.32 $\pm$ 7.51 & 90.22 $\pm$ 3.17 \\
$\pm$SD (\%) & 1000 &  86.14 $\pm$ 0.81 &  91.66 $\pm$ 2.49 &  89.62 $\pm$ 3.10 & 94.26 $\pm$ 0.34 & \textbf{95.06} $\pm$ 0.85 & 94.06 $\pm$ 1.55 \\
 & 2000 & 90.60 $\pm$ 1.82 & 94.00 $\pm$ 1.57 & 94.08 $\pm$ 1.08 & \textbf{96.36} $\pm$ 1.07 & 96.20 $\pm$ 0.95 & \textbf{96.36} $\pm$ 1.59 \\
 & 4000 &  93.20 $\pm$ 0.61 & 97.06 $\pm$ 0.55 & 96.98 $\pm$ 0.48 & \textbf{98.06} $\pm$ 0.52 & 96.84 $\pm$ 1.24 & 97.28 $\pm$ 0.85 \\
   \hline & 125 &  55.33 $\pm$ 12.08 &  57.69 $\pm$ 14.58 &  50.46 $\pm$ 8.81 &  \textbf{80.69} $\pm$  6.56 &   64.75$\pm$ 12.59  &  54.66 $\pm$ 10.23 \\
   & 250 &  78.45 $\pm$ 5.18 &  87.24 $\pm$ 2.69 &  76.49 $\pm$ 8.33 &  \textbf{90.04} $\pm$ 2.43 &  81.89 $\pm$ 2.92 &  81.72 $\pm$ 6.96 \\
  F1$\pm$SD (\%) & 500 &  86.03 $\pm$ 2.91 &  87.78 $\pm$ 3.34 &  88.82 $\pm$ 2.64 &  \textbf{90.51} $\pm$ 2.09 &  82.32 $\pm$ 8.45 &  90.13 $\pm$ 3.21 \\
   & 1000 &     85.88 $\pm$ 0.85 &  91.55 $\pm$ 2.59 & 89.26 $\pm$ 3.35 &  94.23 $\pm$ 0.34 &     \textbf{95.03} $\pm$  0.86 &     94.00 $\pm$ 1.58 \\
   & 2000 &  90.55 $\pm$ 1.80 &  93.97 $\pm$ 1.59 &  94.06 $\pm$ 1.07 &  \textbf{96.36} $\pm$ 1.07 &     96.19 $\pm$ 0.95 &    96.34 $\pm$ 1.61 \\
   & 4000 &  93.18 $\pm$ 0.62 &  97.07 $\pm$ 0.54 &  96.98 $\pm$  0.49 &  \textbf{98.06} $\pm$ 0.52 &     96.83 $\pm$ 1.25 &     97.28 $\pm$ 0.85 \\
 \hline & 125 &  1.33 $\pm$ 0.31 &  2.17 $\pm$ 1.69 &  3.06 $\pm$ 2.43 &  \textbf{1.06} $\pm$ 0.55 &  1.82 $\pm$ 0.60 &  5.87 $\pm$ 2.40 \\
 & 250 &  0.91 $\pm$ 0.32 &  0.71 $\pm$ 0.10 &  1.58 $\pm$ 0.83 &  \textbf{0.42} $\pm$ 0.12 &  0.98 $\pm$ 0.28 &  0.82 $\pm$ 0.50 \\
Loss$\pm$SD & 500 &  0.45 $\pm$ 0.10 &  0.51 $\pm$ 0.19 &  0.45 $\pm$ 0.11 &  \textbf{0.28} $\pm$ 0.08 &  0.65 $\pm$ 0.34 &  0.37 $\pm$ 0.17 \\
 & 1000 &  0.49 $\pm$ 0.06 &  0.32 $\pm$ 0.13 &  0.37 $\pm$ 0.12 &  0.20 $\pm$ 0.02 &  \textbf{0.18} $\pm$ 0.04 &  0.20 $\pm$ 0.05 \\
 & 2000 &  0.31 $\pm$ 0.07 &  0.23 $\pm$ 0.06 &  0.20 $\pm$ 0.05 &  \textbf{0.11} $\pm$ 0.05 &  0.13 $\pm$ 0.03 &  0.13 $\pm$ 0.06 \\
 & 4000 &  0.21 $\pm$ 0.03 &  0.10 $\pm$ 0.02 &  0.12 $\pm$ 0.02 &  \textbf{0.06} $\pm$ 0.01 &  0.12 $\pm$ 0.05 &  0.09 $\pm$ 0.03 \\
\bottomrule
\end{tabular} 
} }
\end{table*}

\section{Results}

We compare two transfer learning methods, ResNet50 and Inception-v3, and four self-supervised learning methods, SwAV, MoCo-v2, SimCLR-v1, and BYOL, on different training set sizes ranging from n=125 to n=4000, under both balanced and imbalanced training set scenarios. Table \ref{tab:table1} depicts the three performance metrics (accuracy, weighted F1, cross entropy loss) of the balanced dataset, while Table \ref{tab:table2} shows the three performance metrics of the imbalanced dataset. We further select dataset sizes of 125, 500, 2000, and 4000 to depict bar plots of average metrics with standard deviation for both balanced and imbalanced scenarios, as shown in the Figure \ref{barplot}.

Followings are the summary of observations we notice based on the Table \ref{tab:table1}, the Table \ref{tab:table2} and Figure \ref{barplot} using accuracy and weighted F1 as the performance metric. Within the two transfer learning methods, ResNet50 consistently outperforms Inception-v3 across all dataset sizes and scenarios. Among the self-supervised learning methods, MoCo-v2 has the best performance when the training set size is small (n= 125, 250) regardless of the balance or imbalance of the training dataset. When the training set size is relatively large (n = 500, 1000, 200, 4000), the MoCo-v2 archives the best performance under the imbalanced scenario, while BYOL surpasses the other self-supervised learning methods under the balanced scenario. Overall, the best self-supervised learning method consistently outperforms the best transfer learning method across all training set sizes and under both balanced and imbalanced scenarios. The performance improvement is more prominent under the imbalanced scenario than the balanced one. Notably, MoCo-v2 outperforms other five models under the imbalanced scenario for all training set sizes (except for dataset size of 1000) and further reduces the variance. MoCo-v2 has the lowest variance in term of accuracy and weighted F1 compared to other models. Interestingly, SwAV has inferior performance compared to the other three self-supervised learning models.

Despite using weighted cross entropy during the training of the imbalanced datasets, we observe that all methods still have worse performance on the imbalanced datasets as compared to the balanced counterparts with the same dataset size. This is even more evident in small datasets (n $<=$ 1000). Although we can obtain the same trends as above using either the accuracy or weighted F1 score as a performance metric, the weighted F1 score is lower than the accuracy when the training is on an imbalanced small dataset (n $<=$ 500). This suggests that the models trained on the small imbalanced datasets are more prone to overfitting and fail to generalize to predict the minority classes. Of note, the MoCo-v2 method greatly reduces this effect as compared to the other five methods. Overall, the MoCov2 has consistently good performance across all training set sizes regardless of balanced and imbalanced scenarios. BYOL outperforming the others for larger dataset sizes in balanced scenarios.

\begin{figure}[h]
\vskip 0.0in
\begin{center}
\centerline{\includegraphics[width=1 \columnwidth]{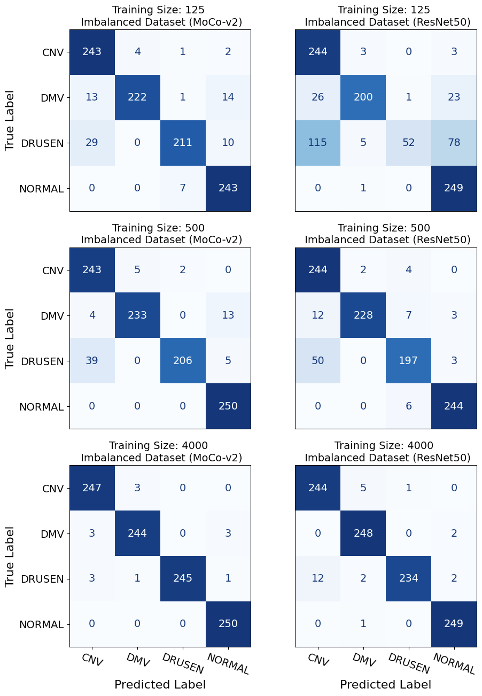}}
\end{center}
\vskip -0.2in
\caption{Confusion matrix for our best results of transfer learning (ResNet50) and self-supervised learning (Moco-v2) pre-trained model in comparison of different training sizes (125, 500 and 4000).}
\label{confusion_matrix}
\end{figure}

\begin{table}[h]
\floatconts
  {tab:table3}%
  {\caption{Performance comparison with other models on the UCSD Dataset}}%
 {%
%  \vskip -0.1in
 \resizebox{1\linewidth}{!}{%
\begin{tabular}{llcc}
\toprule
\toprule
% & & \multicolumn{2}{c}{Transfer Learning} & \multicolumn{4}{c}{Self-Supervised Learning} \\ 
% \cmidrule(lr){3-4} \cmidrule(lr){5-8}
Reference & Model &  Training Size & Accuracy (\%)  \\
  \midrule
 Our $1^{st}$ Result (SSL) & \textbf{SimCLR v1}   & 4,000 &  \textbf{98.84} \\
 Our $2^{nd}$  Result (SSL) & \textbf{MoCo-v2}   & 4,000 &  \textbf{98.62} \\
 Our $1^{st}$ Result (TL) & ResNet50   & 4,000 &  97.96 \\
  \citet{kermany2018identifying}  & Inception-v3 & 4,000 &  93.40 \\
 \midrule
 \citet{sotoudeh2022multi} &  FPN-VGG16 & 108,309  & 98.40 \\
 \citet{paul2020octx} & OCTx & 83,484  & 98.53 \\
 \citet{rastogi2019detection} & DenseNet-C & 108,309   & 97.65 \\
  \citet{kermany2018identifying}  & Inception-v3   & 108,312 &  96.60 \\
\bottomrule
\end{tabular}%
}  }
\end{table}

Table \ref{tab:table3} lists our best results with comparision with other models in the literature. SimCLR-v1 and MoCo-v2 trained on  4000 images with balanced distribution achieve the best and $2^{nd}$ best results with accuracy of 98.84\% and 98.62\%, respectively. Both outperform the Inception-v3  trained on balanced 4000 images in \citet{kermany2018identifying}. Our self-supervised learning with SimCLR v1 schemes also achieve better performance as compared to other deep learning models from previous studies, which are trained at a considerable large scale dataset size. For example, OCTx in  \citet{paul2020octx} is trained on 83,484 OCT images and achieves an accuracy of 98.53\% and FPN-VGG16 in \citet{sotoudeh2022multi} is trained on 108,309 OCT images with accuracy of 98.40\%. The results in Table \ref{tab:table3} indicate that self-supervised learning schemes could be a promising option for training medical images with scarcity, and still achieve comparable accuracy with less computation loads. 

Figure \ref{confusion_matrix} presents the confusion matrices for ResNet50, the best transfer learning model, and MoCo-v2, the best self-supervised learning model with different training set sizes (n = 125, 500,  4000) under the imbalanced training scenario. The confusion matrices show how the model predictions compare to the true labels for each condition with the diagonal elements indicating the true predictions, and the off-diagonal elements indicating the misclassification. We observe that the DRUSEN class has the lowest true prediction, while the DMV class has the second lowest true prediction in the small datasets (n = 125, 500) for both ResNet50 and MoCo-v2. The most common error is that the DRUSEN class gets misclassified as CNV. This occurs more often in the ResNet50 method as compared to MoCo-v2. This supports our previous observations that self-supervised learning methods perform better than transfer learning methods for the imbalanced dataset, and that the model improves as the models are trained on more data from the minority classes.

\section{Discussion}

Automated medical diagnosis is still challenging due to the scarcity of medical images and the expensive labor annotation costs. Integrating deep neural networks with image-based medical diagnosis has risen in popularity. Nevertheless, deep neural networks need to be applied on large scale dataset and are very time consuming in the training stage, limiting the applications in other medical fields with low data availability. Transfer learning has been proposed to tackle the aforementioned limitation. Self-supervised learning is a good alternative to transfer learning and has been shown comparable performance to its supervised counterpart for classification from an imbalanced medical image dataset.

In this study, we assess four pretrained self-supervised learning models and two transfer learning models using small-scale OCT images ranging from 125 to 4000 with balanced or imbalanced distribution for training. The self-supervised learning with MoCo-v2 and SimCLR-v1 scheme achieve state-of-art performance on the UCSD Dataset for retinal diseases classification using only 4,000 OCT images as compared to other studies that achieve similar performance using 108,309 OCT images. Of note, the SwAV has consistently inferior performance in various training set sizes as compared to the other three self-supervised models, which suggests self-supervised learning via instance-based discrimination is the better option than cluster-based discrimination in OCT images classification. Furthermore, the proposed self-supervised learning framework is computational efficiency and takes very short time to complete one training task. It takes only about one hour for 1-fold training with 4,000 OCT images as training data using PyTorch, powered by the NVIDIA T4 GPU in Google Colab service, encouraging the promising usage of self-supervised learning models for future automated medical diagnosis. We also envision that this framework could be further generalized and hopefully increase availability of automated medical diagnosis models for other treatable diseases which have low training medical images.

There are, however, two observable constraints here. Firstly, the performance is still confined when the dataset size is extremely small (around 125). Second, although high accuracy in classification for all classes (CNV, DME, Drusen, Normal), the most common error is that the Drusen class gets misclassified as CNV, which also corresponds to actual medical cases since Drusen is the early stage of CNV. However, given our problem statement, it is favorable to have false positives which might require further checks, rather than missing cases which needed urgent referral but got classified as observation-only. In the future, we will further evaluate this self-supervised learning framework on other small scale and unlabeled image dataset. Moreover, the self-supervised learning models used in this study are pre-trained on the ImageNet. In the future, pre-trained models can be trained on medical images, such as chest-x-ray or OCT images, which could learn domain-relevant knowledge and potentially further improve the performance in very small-scale datasets. 

\section{Conclusion }

In this study, we exploit pre-trained transfer learning and self-supervised learning models on UCSD Dataset in classification of treatable retinal diseases. We applied four pre-trained self-supervised models and two transfer learning models in different scales of sub-dataset size, ranging from 125 to 4000. The self-supervised learning models, MoCo-v2 and SimCLR-v1, achieves the state-of-art accuracy of 98.62\% and 98.84\% using only 4,000 training images, respectively. Our experiment results suggest pretrained self-supervised learning models outperform the transfer learning models when using small-scale training data under both the balanced and imbalanced training scenarios. MoCo-v2 has consistent good performance under the imbalanced scenario and is especially powerful when the training set is less than 500 images. This demonstrates the advantage of self-supervised learning models in OCT images classification under very limited training images and data imbalance. Our experiment results provide guidance for training medical diagnosis tools with low data availability.

\appendix

\end{document}